\documentclass[conference]{IEEEtran}
\IEEEoverridecommandlockouts

\usepackage{cite}
\usepackage{amsmath,amssymb,amsfonts}
\usepackage{bm}
\usepackage{algorithmic}
\usepackage{graphicx}
\usepackage{textcomp}
\usepackage{xcolor}
\usepackage{booktabs}
\usepackage{multirow}
\usepackage[caption=false]{subfig}
\usepackage{url}

\def\BibTeX{{\rm B\kern-.05em{\sc i\kern-.025em b}\kern-.08em
T\kern-.1667em\lower.7ex\hbox{E}\kern-.125emX}}

\begin{document}

\title{Assessing Post-Reform Changes in Risk Disclosure Quality with a Multidimensional Text Analysis Approach}

\author{\IEEEauthorblockN{Nobuhiro Aikawa}
\IEEEauthorblockA{\textit{Degree Programs in Business Sciences} \\
\textit{University of Tsukuba}\\
Tokyo, Japan \\
s2540051@u.tsukuba.ac.jp}
\and
\IEEEauthorblockN{Mitsuo Yoshida}
\IEEEauthorblockA{\textit{Institute of Business Sciences} \\
\textit{University of Tsukuba}\\
Tokyo, Japan \\
mitsuo@gssm.otsuka.tsukuba.ac.jp}
}

\maketitle

\begin{abstract}
While corporate narrative disclosures provide crucial information to capital markets, comprehensively evaluating their qualitative changes over time remains challenging.
Narrative text is inherently multidimensional, meaning that an improvement in one textual dimension often occurs alongside changes in others.
To capture these underlying dynamics, we propose a longitudinal text analysis approach combining Japanese-language NLP metric extraction with paired testing, shift function analysis, and inter-metric correlation.
Our framework extends prior indicator sets by incorporating a \textit{cross-section relevance} indicator to measure topical alignment between risk disclosures and management strategies.
Applying this approach to evaluate Japan's 2019 disclosure reforms, we analyze 19{,}770 firm-year observations over a 10-year period (FY2015--FY2024).
The joint analysis reveals complex shifts in disclosure patterns that are frequently masked by conventional single-indicator methods.
Specifically, we find that while disclosure volume increased substantially, it was accompanied by a decline in readability.
Furthermore, although the overall information structure improved, specific descriptive quality stagnated, and the degree of adaptation varied across market segments.
\end{abstract}


\section{Introduction}

Corporate narrative disclosures, such as risk descriptions and management discussions, constitute a major source of information for capital market participants.
While evaluating the quality of such long-form text is crucial, analyzing textual dimensions in isolation can mask important dynamics.
For example, an increase in disclosure volume might appear as an improvement, but it could be accompanied by a decline in readability or persistent formulaic descriptions at the sentence level.
Capturing these multifaceted changes requires evaluating multiple textual dimensions simultaneously.

Although prior studies have made progress by developing multidimensional indicator systems, they often analyze each metric independently and focus primarily on average effects.
This approach leaves potential inter-metric asymmetries and distributional heterogeneity across firms largely unexplored.
Furthermore, existing frameworks focus on properties internal to a single narrative section, often missing the topical alignment across different sections that recent disclosure reforms increasingly emphasize.

To address these limitations, we present a longitudinal text analysis approach that combines six complementary indicators: volume, specificity, readability, boilerplate, stickiness, and a \textit{cross-section relevance} metric.
This relevance indicator specifically captures the topical alignment between risk disclosures and broader management strategies.
We extract these metrics using Japanese NLP tooling (GiNZA~\cite{Matsuda2020}) and integrate them with three statistical components, paired $t$-tests, shift function analysis~\cite{doksum1974, rousselet2017}, and inter-metric correlation analysis, to examine mean-level shifts, distributional heterogeneity, and inter-metric relationships in a unified manner.

We illustrate this approach through an empirical application to Japan's 2019 regulatory reforms of annual securities reports, analyzing 19{,}770 firm-year observations from 1{,}977 listed companies over a 10-year period (FY2015--FY2024).
More than five years after the Financial Services Agency implemented comprehensive reforms~\cite{FSA2019, FSA2019b} calling for specific and strategy-aligned risk descriptions, the nature of the actual changes in corporate disclosure practices remains debated.
Our analysis reveals three distinct dynamics: a volume--readability trade-off, a structural--descriptive asymmetry, and a market-segment disparity under a uniform regulatory regime. These findings demonstrate how our combined approach can detect patterns difficult to observe with conventional methods.

\section{Related Work}

In this section, we review prior quantitative research on corporate narrative disclosures, outline the background of Japan's 2019 regulatory reforms, and identify gaps in the existing literature.

\subsection{Quantification of Narrative Disclosures}

Prior quantitative research on corporate narrative disclosures has predominantly focused on the U.S.\ context.
Li~\cite{Li2008} applied the Fog Index to 10-K filings and showed that less readable annual reports are associated with lower and less persistent earnings, establishing readability as a measurable and economically meaningful dimension of disclosure quality.
Loughran and McDonald~\cite{Loughran2011} constructed a finance-specific sentiment dictionary and demonstrated that the tone of narrative disclosures is significantly associated with market reactions.
For risk disclosures specifically, Campbell et al.~\cite{Campbell2014} showed that mandatory risk factor disclosures are associated with reduced information asymmetry between firms and investors.
Furthermore, Hope et al.~\cite{Hope2016} demonstrated that \textit{specific} descriptions, operationalized as the proportion of named entities addressing firm-specific circumstances, are associated with lower market uncertainty, underscoring the importance of quality beyond mere volume.

Building on these single-dimension studies, Dyer et al.~\cite{Dyer2017} developed a comprehensive multidimensional indicator system encompassing length, boilerplate, specificity, redundancy, and stickiness.
Applying this system to a long-term analysis of U.S.\ 10-K filings, they revealed that while disclosure volume grows annually, content becomes increasingly boilerplate through year-over-year carryover, accompanied by declines in specificity and readability.
Their multidimensional framework provides the direct conceptual foundation for the indicator set used in this study.
In the Japanese context, Ito et al.~\cite{Ito2021} computed multidimensional indicators, including readability, specificity, boilerplate, and stickiness, for MD\&A and risk information from 2004 to 2018, establishing a baseline for Japanese disclosure analysis.

However, while these studies employ multiple indicators, they typically report each metric independently and focus on mean-level comparisons.
Consequently, the interactions and trade-offs among indicators, as well as the distributional heterogeneity of firm responses, remain largely unexplored.
Moreover, existing indicator sets focus on properties internal to a single narrative section and do not capture topical alignment \textit{across sections}, a property increasingly emphasized by recent disclosure reforms.
To address distributional changes, shift function analysis~\cite{doksum1974}, a nonparametric method for comparing distributions at the quantile level, offers a robust solution.
Although widely applied in psychology and neuroscience~\cite{rousselet2017}, it has seen limited use in corporate disclosure research.

Beyond firm-level characteristics and analytical methodologies, institutional factors such as market listing segments also play a crucial role in shaping disclosure behavior.
For instance, Miihkinen~\cite{Miihkinen2012} examined Finnish listed companies and found that listing on a major market is associated with higher levels of risk disclosure.
This finding highlights that market characteristics significantly influence corporate disclosure practices, motivating our cross-segment analysis within the Japanese regulatory jurisdiction.

\subsection{Empirical Studies on Japan's 2019 Reforms}
Prior to 2019, Japanese risk disclosures were widely criticized for generic content, year-over-year carryover unresponsive to changing business environments, and a lack of coherence with management strategy\cite{FSA2018}. 
To address these issues, the Financial Services Agency revised the disclosure ordinance in 2019, requiring firms to provide specific, plain-language risk descriptions coherent with their business strategies through a principles-based approach. 
Several studies have directly examined changes in narrative disclosures following these reforms.
Nakano and Yuasa~\cite{Nakano2022} analyzed character counts and stickiness in MD\&A sections before and after the reforms, demonstrating that while character counts increased during the initial year of implementation, stickiness also underwent significant changes.
However, such post-reform verifications typically focus on the immediate aftermath of the regulatory shift, leaving medium- to long-term trends largely unexplored.

\subsection{Research Gaps and Contributions}

Despite the accumulated work reviewed above, several critical gaps remain in the literature.
Most notably, medium- to long-term trends following the 2019 reforms have received little attention, and the underlying dynamics among multiple textual dimensions have not been systematically examined.
Furthermore, the distributional heterogeneity of firm responses remains largely unaddressed within a unified analytical framework.

To address these gaps, this study makes three primary contributions.
First, we extend existing indicator sets by incorporating a \textit{cross-section relevance} indicator to measure the topical alignment between risk disclosures and management strategy narratives.
While existing multidimensional frameworks~\cite{Dyer2017, Ito2021} analyze each narrative section independently and do not measure the topical alignment between distinct sections, our proposed metric explicitly captures the inter-section integration targeted by Japan's 2019 reforms.
Second, we adapt the multidimensional indicator framework for Japanese-language corpora.
This includes implementing character-based volume measurement appropriate for a language lacking explicit word boundaries, applying a Japanese-specific readability formula~\cite{Lee2016}, and extending the NER taxonomy of Hope et al.~\cite{Hope2016} for risk contexts.
Third, we integrate NLP-based metric extraction with a comprehensive statistical toolkit encompassing paired testing, shift function analysis, and inter-metric correlation.
This unified approach allows us to examine mean-level shifts, distributional heterogeneity, and inter-metric relationships simultaneously.
We demonstrate the efficacy of these contributions through a long-term empirical application to the 2019 Japanese disclosure reforms.

\section{Methodology}

This section outlines our proposed analytical framework.
We detail the procedures for extracting six multidimensional indicators from Japanese text and describe the statistical toolkit used to systematically evaluate qualitative changes.

\subsection{Approach Overview}

Our analytical framework integrates two primary modules.
First, a multidimensional metric extraction module quantifies six complementary dimensions of narrative text.
Second, a statistical analysis module combines mean-level testing, distributional comparisons, and inter-metric correlation analysis to systematically identify shifts in disclosure quality.
By jointly analyzing these indicators, we can capture the nuanced dynamics of disclosure quality that single-metric evaluations often obscure.

\subsection{Definition and Measurement of Metrics}

We operationalize disclosure quality through the six metrics detailed below.
All morphological analysis and named entity recognition are performed using GiNZA~\cite{Matsuda2020}, a Japanese NLP library built on a Transformer architecture (ja-ginza-electra) that provides context-dependent representations learned from large-scale corpora.

\subsubsection{\textbf{Volume}}

We measure disclosure volume using the total character count of the risk information section.
Because Japanese text lacks explicit word boundaries, character counts provide a more objective and robust proxy for textual volume than word approximations.
To normalize the distribution and mitigate the influence of extreme outliers, we apply a logarithmic transformation to the raw character count ($N_{\text{char}}$), excluding whitespace:
\begin{equation}
\text{Volume} = \ln(1 + N_{\text{char}})
\label{eq:volume}
\end{equation}

\subsubsection{\textbf{Specificity}}

Building on Hope et al.~\cite{Hope2016}, we operationalize specificity as the density of concrete named entities within the narrative.
Using GiNZA's named entity recognizer, we extract entities across four risk-relevant categories: organizational entities (e.g., companies, governments,
and their key personnel such as founders and executives), locations and facilities, quantities and temporal expressions, and business-environment entities (e.g., products, laws, natural disasters).\footnote{Named entity recognition was performed using ja-ginza-electra, which supports fine-grained multi-class entity extraction based on the Extended Named Entity Hierarchy and OntoNotes 5.0 taxonomy. This four-category selection extends Hope et al.~\cite{Hope2016}'s operationalization by subsuming persons under the organizational category and adding business-environment entities.}
Specificity is computed as the ratio of these retained named entity tokens to the total number of content words (nouns, verbs, adjectives, and adverbs):
\begin{equation}
\text{Specificity} = \frac{\text{Count(Named Entity Tokens)}} {\text{Count(Content Words)}}
\label{eq:specificity}
\end{equation}
Higher scores indicate a shift away from generic abstractions toward concrete, evidence-based descriptions.

\subsubsection{\textbf{Readability}}

To assess linguistic complexity, we employ the Japanese readability formula developed by Lee~\cite{Lee2016}, which is specifically calibrated for Japanese language comprehension:
\begin{equation}
\begin{split}
\text{Readability} &= 11.724 - 0.056 \times \text{MeanSentLen} \\ &\quad - 0.126 \times \text{SinoRatio} - 0.042 \times \text{NativeRatio} \\ &\quad - 0.145 \times \text{VerbRatio} - 0.044 \times \text{ParticleRatio}
\end{split}
\label{eq:readability}
\end{equation}
Intuitively, the formula captures the notion that texts with longer sentences and higher proportions of Sino-Japanese words or verbs are more difficult to comprehend.
Sino-Japanese and native-Japanese words are identified using the Japanese Language Education Vocabulary List maintained by the Japanese Learning Dictionary Support Group\footnote{\url{http://jhlee.sakura.ne.jp/JEV/}}.
A higher score indicates greater readability.

\subsubsection{\textbf{Boilerplate}}

Following Ito et al.~\cite{Ito2021}, we measure the prevalence of boilerplate language by identifying highly standardized phrasing shared across firms.
Specifically, we define a ``boilerplate phrase'' as an 8-word sequence appearing in at least 30\% of all corporate disclosures within a given fiscal year.\footnote{These thresholds follow Ito et al.~\cite{Ito2021}, who apply an 8-gram, 30\% threshold to Japanese disclosures. This contrasts with Dyer et al.~\cite{Dyer2017}'s 4-gram, 75\% threshold for boilerplate in English 10-K filings, likely reflecting the finer granularity of Japanese morphological tokens.}
Any sentence containing such a phrase is classified as formulaic.
The boilerplate metric is then calculated as the word count ratio of these formulaic sentences to the entire document:
\begin{equation}
\text{Boilerplate} = \frac{\sum_{s \in S_{\text{BP}}} |s|} {\sum_{s \in S_{\text{all}}} |s|}
\label{eq:boilerplate}
\end{equation}
Here, $S_{\text{all}}$ represents the set of all sentences in the document, $S_{\text{BP}}$ represents the set of formulaic sentences, and $|s|$ represents the word count of sentence $s$.
A higher score indicates a heavier reliance on generic, standardized descriptions shared by other companies.

\subsubsection{\textbf{Stickiness}}

To capture the persistence of narrative content over time, we measure stickiness based on the methodology of Carl\'{e} et al.~\cite{carle2023text}.
Japanese risk disclosures are frequently drafted by incrementally revising the previous year's text rather than writing anew.
While character-level edit distance (Levenshtein distance~\cite{Levenshtein1966}) is well-suited to measure these precise updates, applying it naively at the document level inflates penalties when sentences are merely reordered.
To resolve this, we segment documents into sentences and apply the Hungarian algorithm~\cite{Kuhn1955} to find the optimal sentence-level alignment that minimizes the total edit distance.
Unmatched sentences, those entirely added in year $t$ or removed from year $t-1$, incur their full character counts as an edit cost.
The stickiness score is derived by subtracting this normalized edit distance from 1:
{\small
\begin{equation}
\text{Stickiness} = 1 - \frac{\displaystyle \sum_{(i,j) \in \mathcal{M}} d(s_i, r_j) + \sum_{i \in \mathcal{U}_t} |s_i| + \sum_{j \in \mathcal{U}_{t-1}} |r_j|}{\displaystyle \sum_{k=1}^{n} |s_k| + \sum_{l=1}^{m} |r_l|}
\label{eq:stickiness}
\end{equation}
}%
In this equation, $d(s_i, r_j)$ represents the edit distance between a matched sentence pair, $|s_i|$ and $|r_j|$ represent the character counts of the respective sentences, $\mathcal{M}$ is the set of optimal pairs identified by the algorithm, and $\mathcal{U}$ denotes the set of unmatched sentences.
A score closer to 1 indicates a higher degree of similarity to the previous year's disclosure.

\begin{table*}[tp]
\centering \caption{Results of Mean Difference Tests and Effect Sizes.}
\label{tab:results}
\begin{tabular}{lrrrrrr}
\toprule & \multicolumn{2}{c}{Pre-Reform} & \multicolumn{2}{c}{Post-Reform} & & \\ \cmidrule(lr){2-3} \cmidrule(lr){4-5} Metric & \multicolumn{1}{c}{Mean} & \multicolumn{1}{c}{SD} & \multicolumn{1}{c}{Mean} & \multicolumn{1}{c}{SD} & \multicolumn{1}{c}{Diff} & \multicolumn{1}{c}{Glass's $\Delta$} \\ \midrule Volume & 7.655 & 0.709 & 8.145 & 0.628 & $0.490^{***}$ & 0.691 \\ Specificity & 0.055 & 0.044 & 0.049 & 0.031 & $-0.006^{***}$ & $-0.129$ \\ Readability & 7.946 & 0.737 & 7.766 & 0.731 & $-0.180^{***}$ & $-0.244$ \\ Boilerplate & 0.329 & 0.177 & 0.328 & 0.177 & $-0.001^{***}$ & $-0.005$ \\ Stickiness & 0.957 & 0.053 & 0.922 & 0.066 & $-0.035^{***}$ & $-0.655$ \\ Relevance & 0.213 & 0.121 & 0.268 & 0.119 & $0.054^{***}$ & 0.448 \\ \bottomrule
\end{tabular}
\begin{flushleft}
\footnotesize \textbf{Note}: This table compares the mean values of metrics before and after the reform, showing quantitative expansion alongside weaker descriptive-quality metrics.
Pre-reform period is FY2015--FY2018; post-reform period is FY2020--FY2024.
Stickiness is calculated for FY2016--FY2018 (pre) and FY2020--FY2024 (post).
Diff indicates the difference in mean values before and after the reform using paired $t$-tests.
The Boilerplate effect size (Glass's $\Delta=-0.005$) indicates a substantively negligible change despite statistical significance.
$^{***}p<0.01$, $^{**}p<0.05$, $^{*}p<0.1$.
\end{flushleft}
\end{table*}

\subsubsection{\textbf{Relevance}}

We define relevance as the lexical-topical similarity between the risk information section and the management strategy section (i.e., ``Management Policy, Business Environment, and Issues to Address'').
This cross-section indicator represents a structural departure from within-section metrics like stickiness.
While stickiness evaluates year-over-year character revisions within the same section, relevance assesses the topical alignment between two distinct sections within the same document.
Consequently, we employ a bag-of-words approach with TF-IDF weighting, which properly emphasizes shared content-bearing keywords while penalizing generic vocabulary.
We compute the cosine similarity between the L2-normalized TF-IDF document vectors:
\begin{equation}
\text{Relevance} = \frac{\mathbf{v}_{\text{risk}} \cdot \mathbf{v}_{\text{ref}}}{\|\mathbf{v}_{\text{risk}}\| \|\mathbf{v}_{\text{ref}}\|} = \sum_{i=1}^{n} w_{i,\text{risk}} \cdot w_{i,\text{ref}}
\label{eq:relevance}
\end{equation}
Here, $\mathbf{v}_{\text{risk}}$ and $\mathbf{v}_{\text{ref}}$ represent the document vectors for the risk and management strategy sections, respectively, and $w_i$ is the TF-IDF weight of word $i$.
A score closer to 1 indicates stronger topical overlap between the disclosed risks and the firm's strategic focus.
Notably, this cross-section relevance metric addresses the inter-section alignment often missing from prior multidimensional frameworks~\cite{Dyer2017, Ito2021}.

\subsection{Statistical Analysis Methods}

Our statistical module employs three complementary methods to unpack the dynamics of disclosure quality.

\textbf{Paired $t$-tests and effect sizes.}
To identify mean-level shifts following the regulatory reforms, we conduct paired $t$-tests on a balanced panel of 1{,}977 firms tracked over a decade.
By analyzing within-firm differences, we directly compare each company's pre- and post-reform disclosure behavior, calculating Glass's $\Delta$ to determine the substantive magnitude of these shifts.

\textbf{Shift function analysis.}
While paired $t$-tests establish whether the mean of within-firm differences deviates from zero, they obscure whether these changes were uniform or driven by specific firm quantiles.
To capture this distributional heterogeneity, we employ shift function analysis~\cite{doksum1974}.
Following Rousselet et al.~\cite{rousselet2017}, we compute decile-level shift functions with 95\% percentile bootstrap confidence intervals based on 500 resamples of within-firm differences at each decile, precisely visualizing where in the distribution the most substantial transformations occurred.

\textbf{Inter-metric correlation analysis.}
Finally, to expose potential relationships between quantitative expansion and qualitative transformation, we analyze the inter-metric correlations among the rates of change for all six indicators.
This step is critical for revealing multidimensional dynamics that remain invisible when indicators are examined in isolation.

\begin{figure*}[tp]
  \centering
  \begin{minipage}{0.48\textwidth}
    \centering
    \subfloat[Volume\label{fig:ts_volume}]{\includegraphics[width=0.48\linewidth]{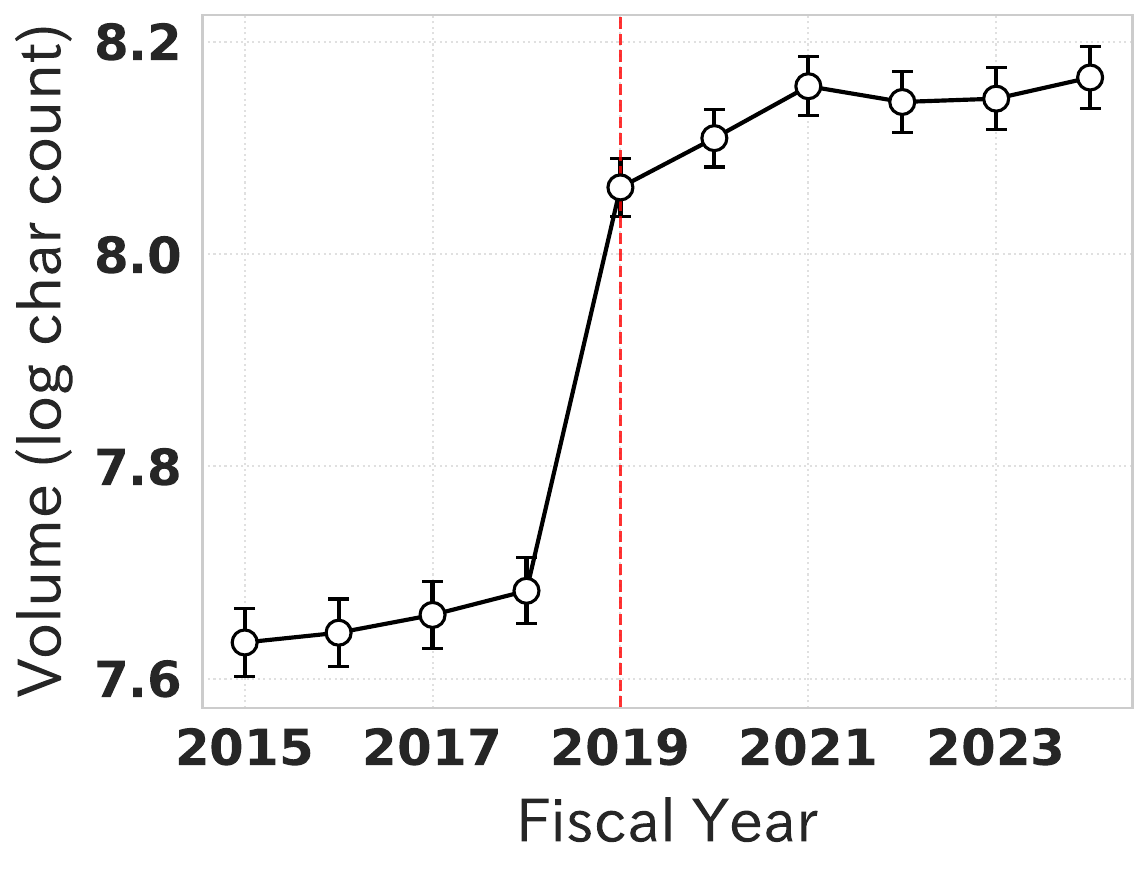}}%
    \hfill
    \subfloat[Specificity\label{fig:ts_specificity}]{\includegraphics[width=0.48\linewidth]{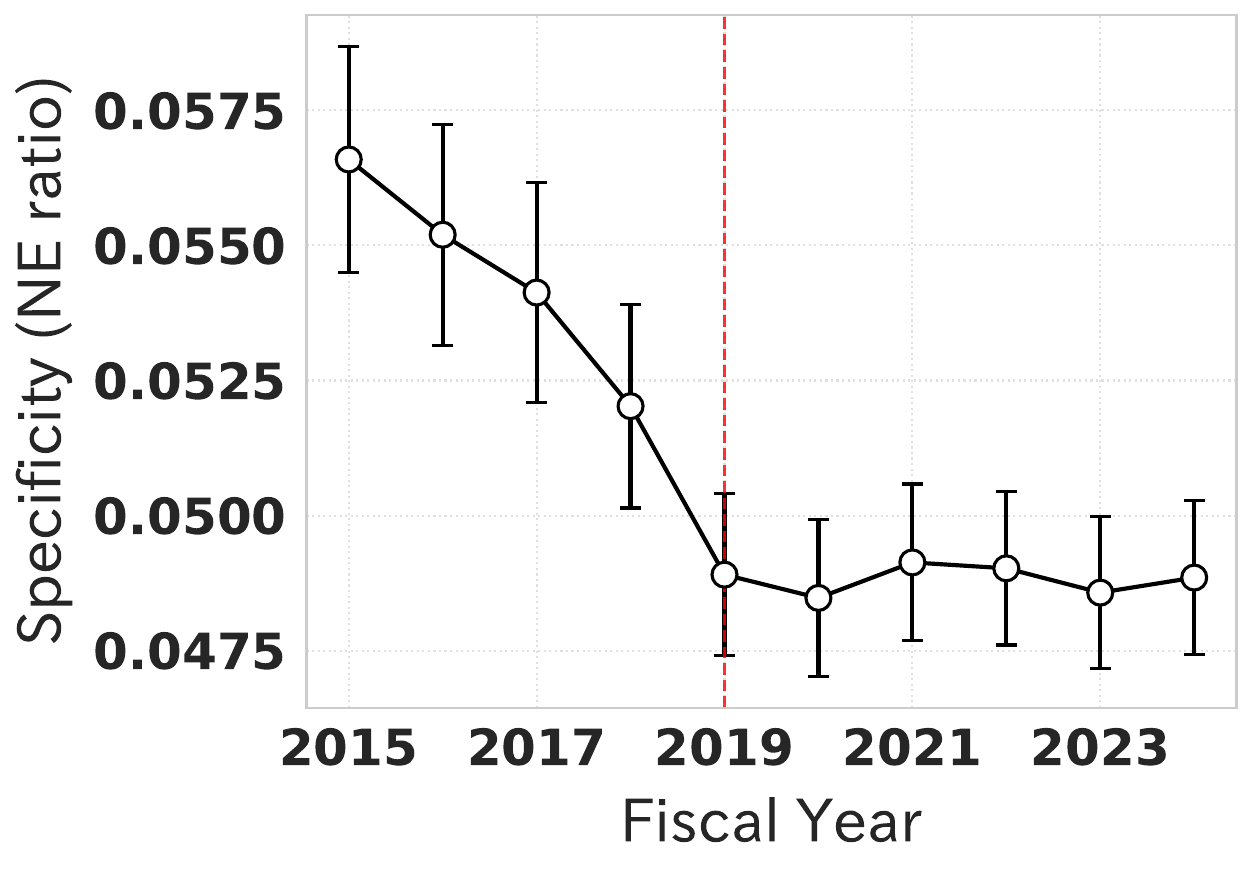}}\\
    \subfloat[Readability\label{fig:ts_readability}]{\includegraphics[width=0.48\linewidth]{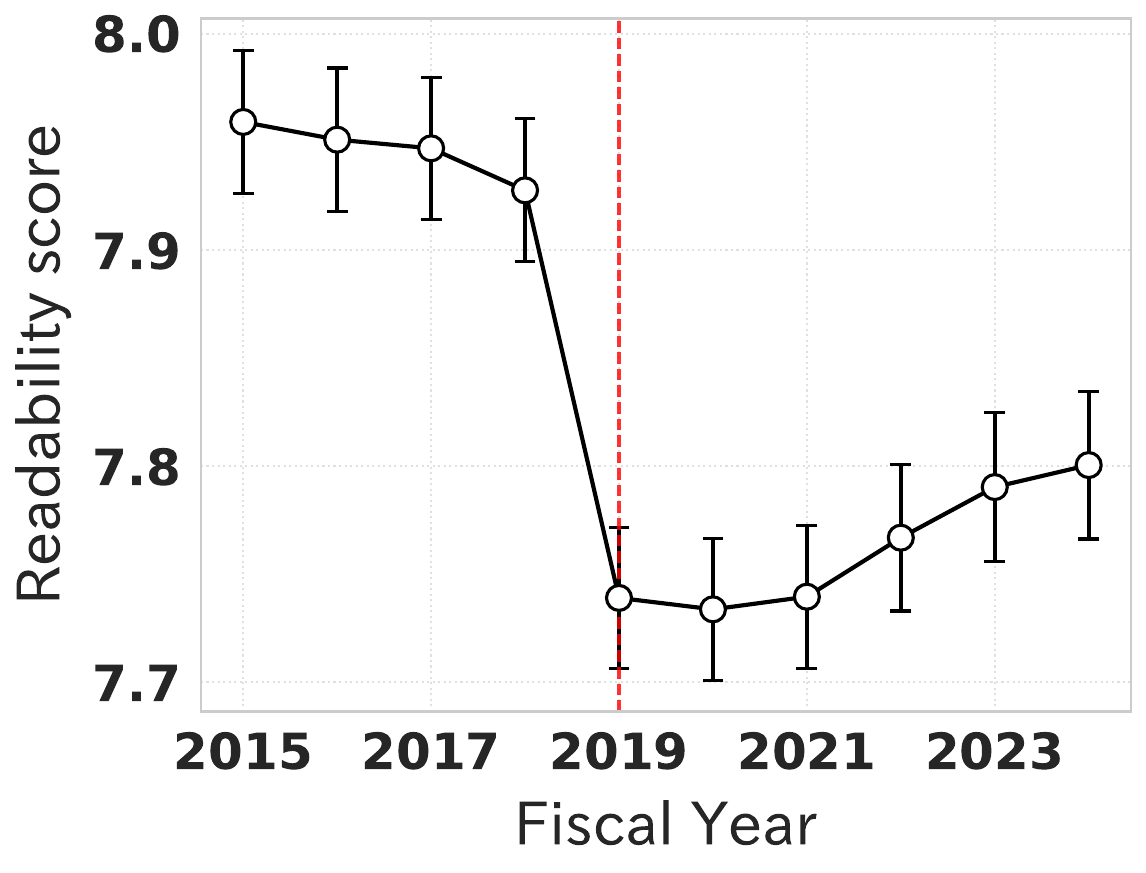}}%
    \hfill
    \subfloat[Boilerplate\label{fig:ts_boilerplate}]{\includegraphics[width=0.48\linewidth]{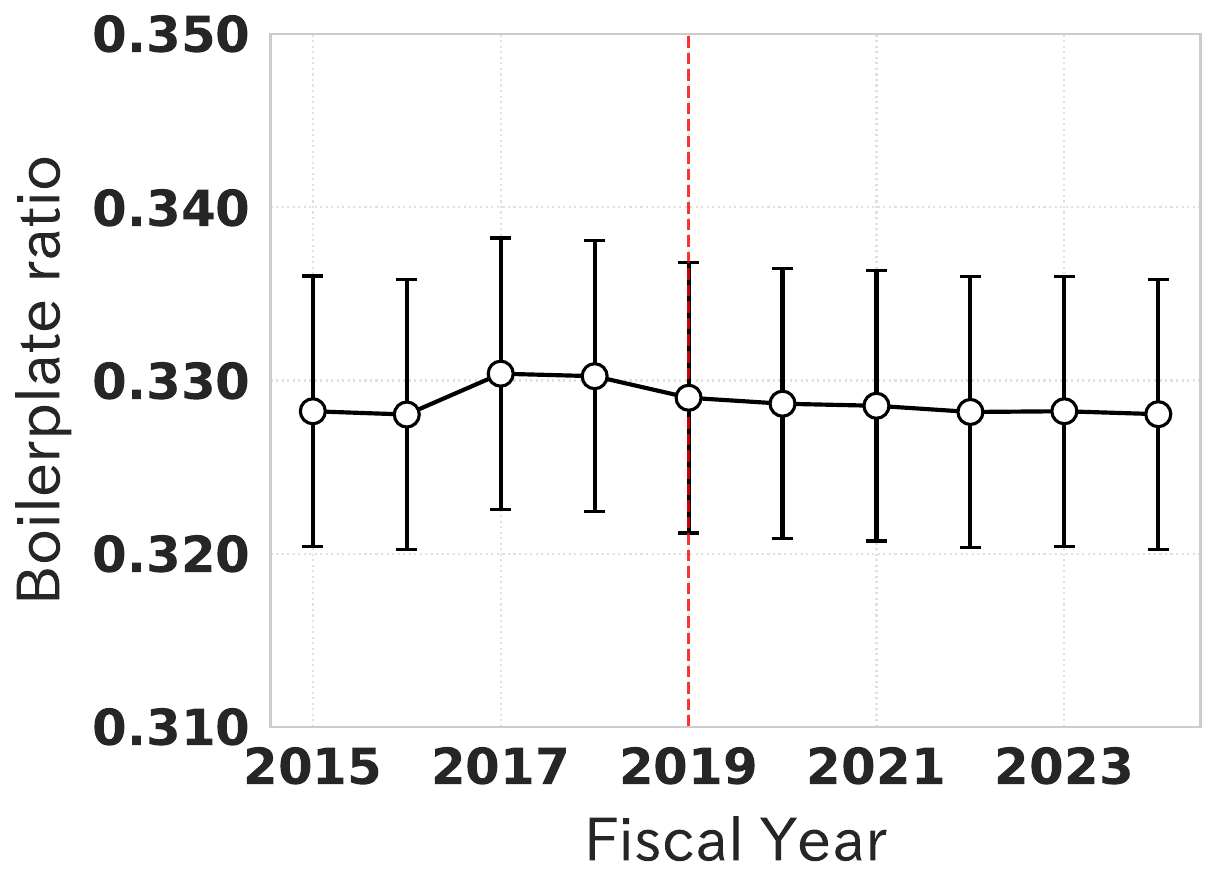}}\\
    \subfloat[Stickiness\label{fig:ts_stickiness}]{\includegraphics[width=0.48\linewidth]{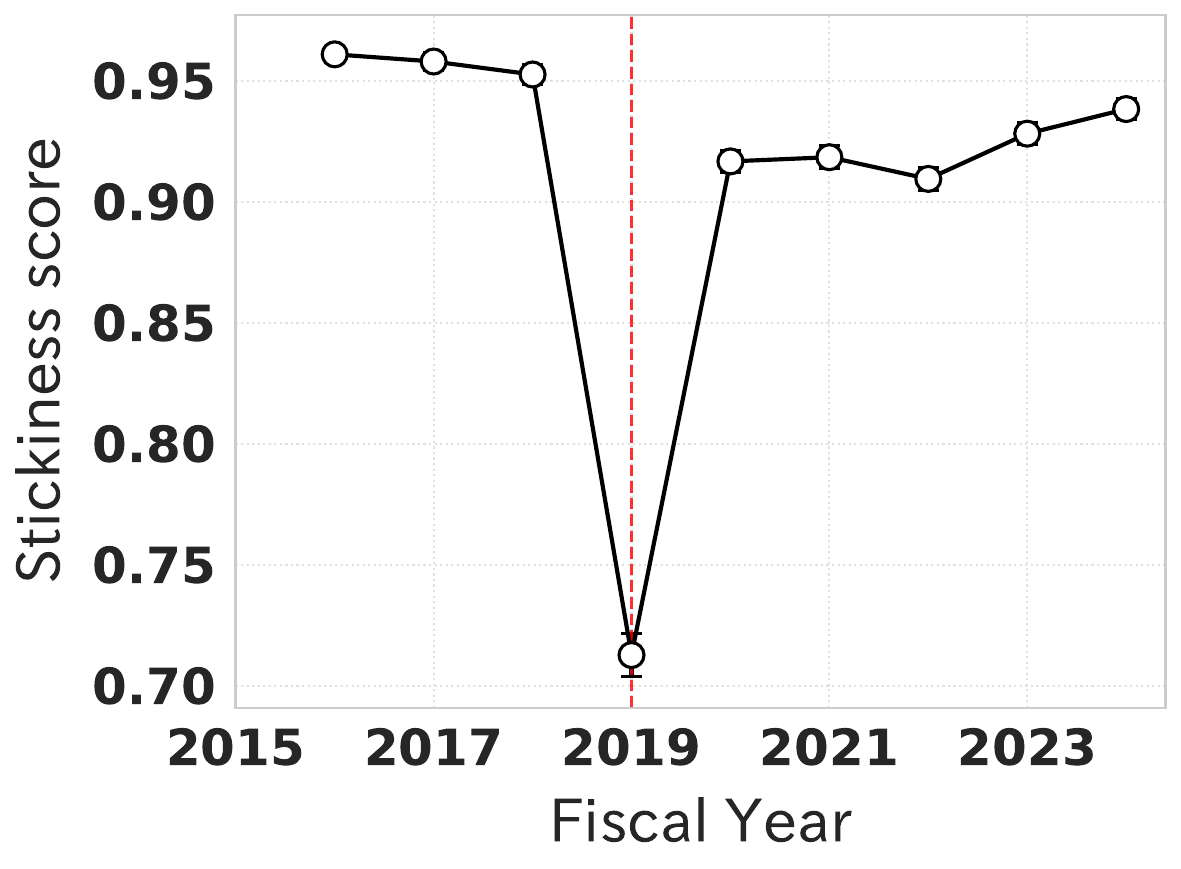}}%
    \hfill
    \subfloat[Relevance\label{fig:ts_relevance}]{\includegraphics[width=0.48\linewidth]{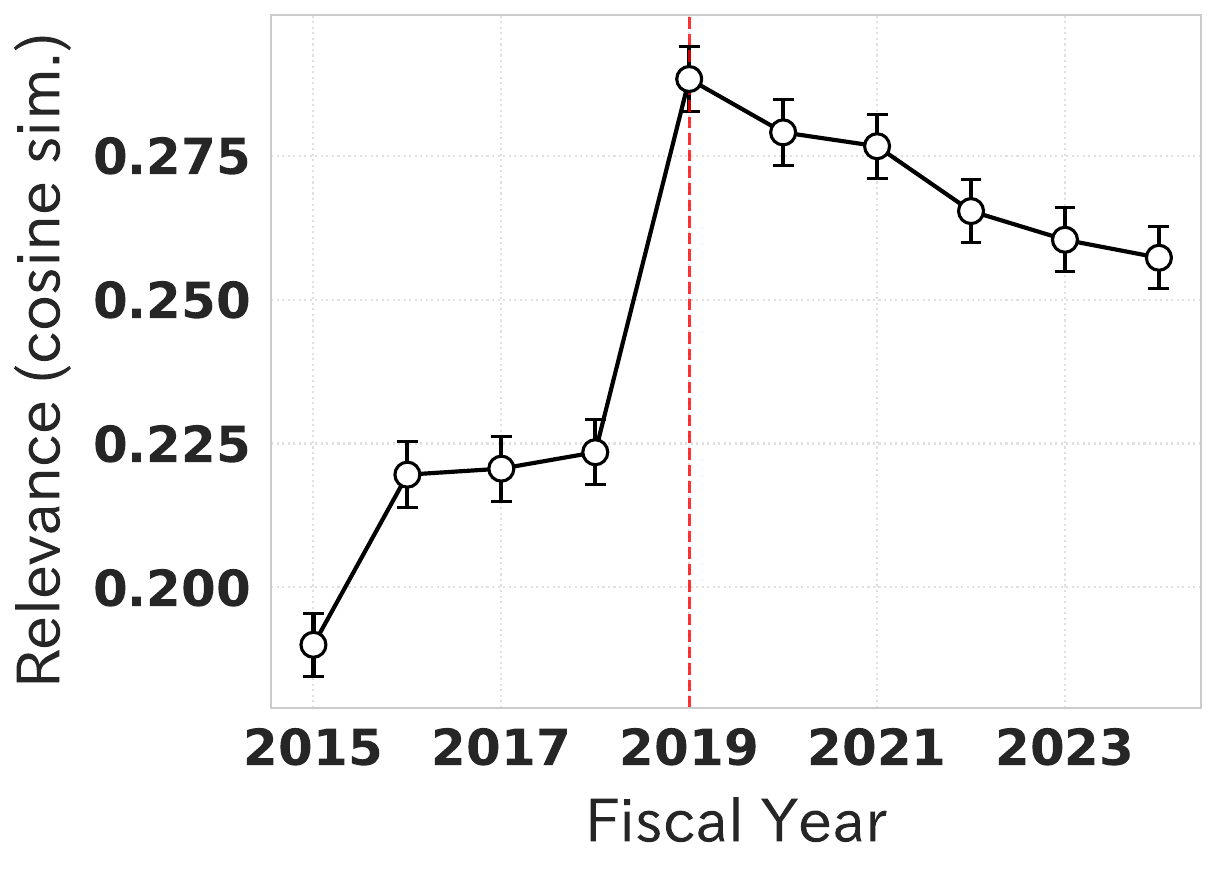}}
    \caption{Time-series trends of quantification metrics.
    The red dashed line indicates the implementation of the 2019 reform (FY2019), and error bars show 95\% confidence intervals.
    The figure shows that volume and relevance increased and stickiness decreased after FY2019, while specificity and readability moved downward.}
    \label{fig:timeseries}
  \end{minipage}%
  \hfill
  \begin{minipage}{0.48\textwidth}
    \centering
    \subfloat[Volume\label{fig:shift_volume}]{\includegraphics[width=0.48\linewidth]{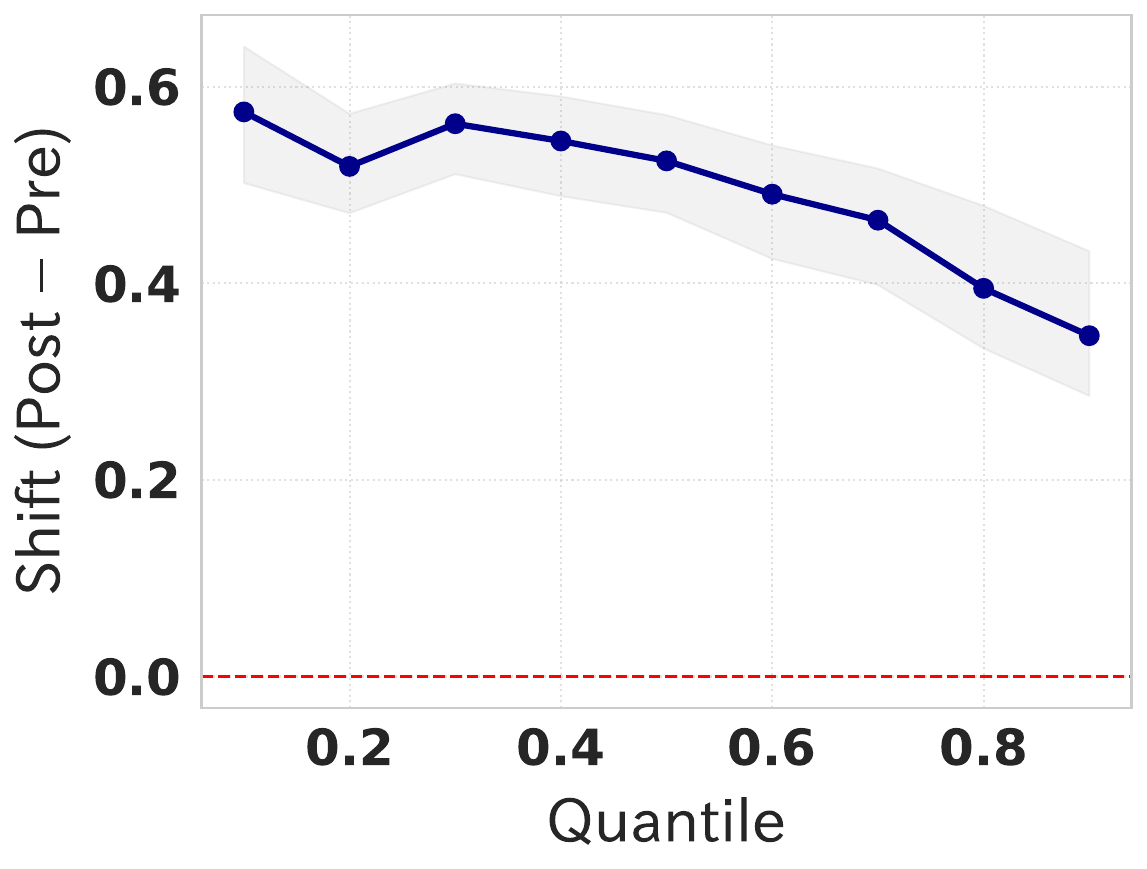}}%
    \hfill
    \subfloat[Specificity\label{fig:shift_specificity}]{\includegraphics[width=0.48\linewidth]{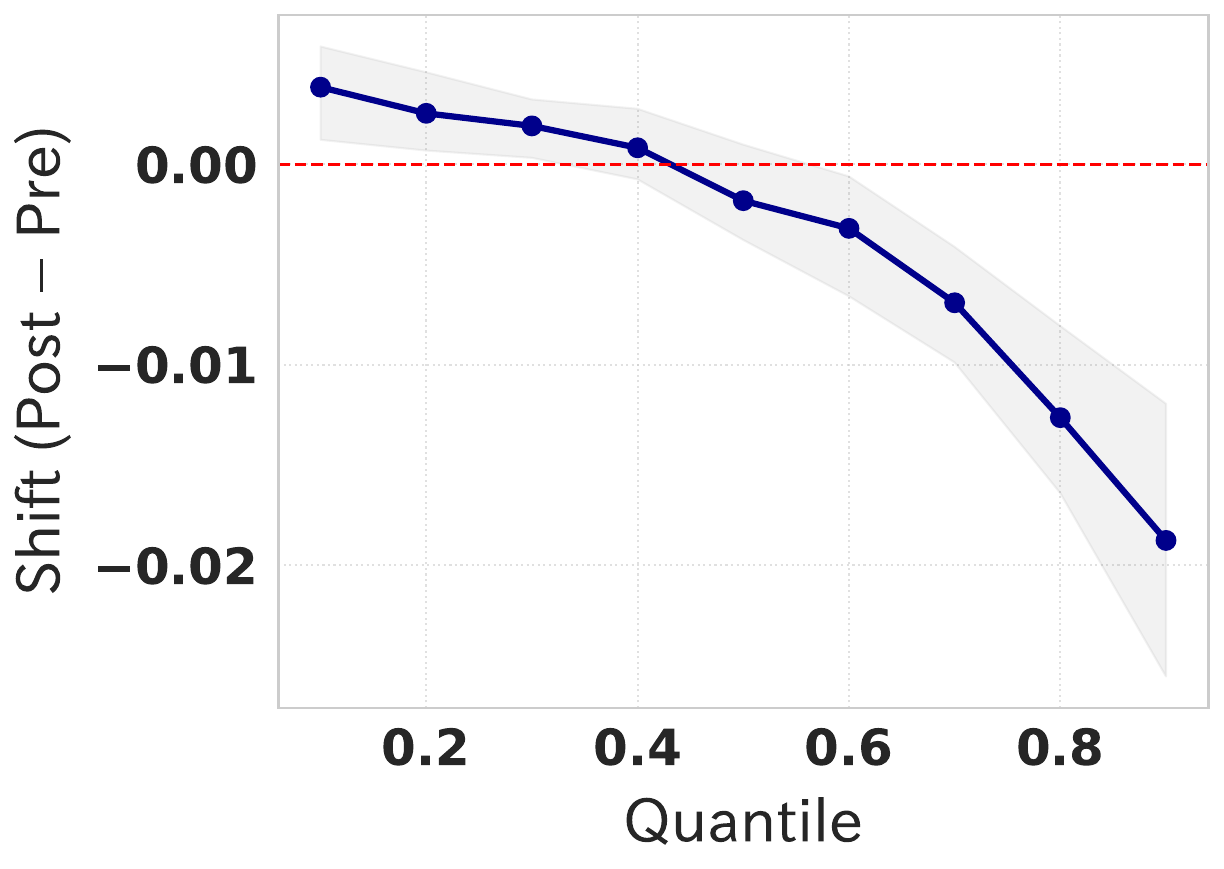}}\\
    \subfloat[Readability\label{fig:shift_readability}]{\includegraphics[width=0.48\linewidth]{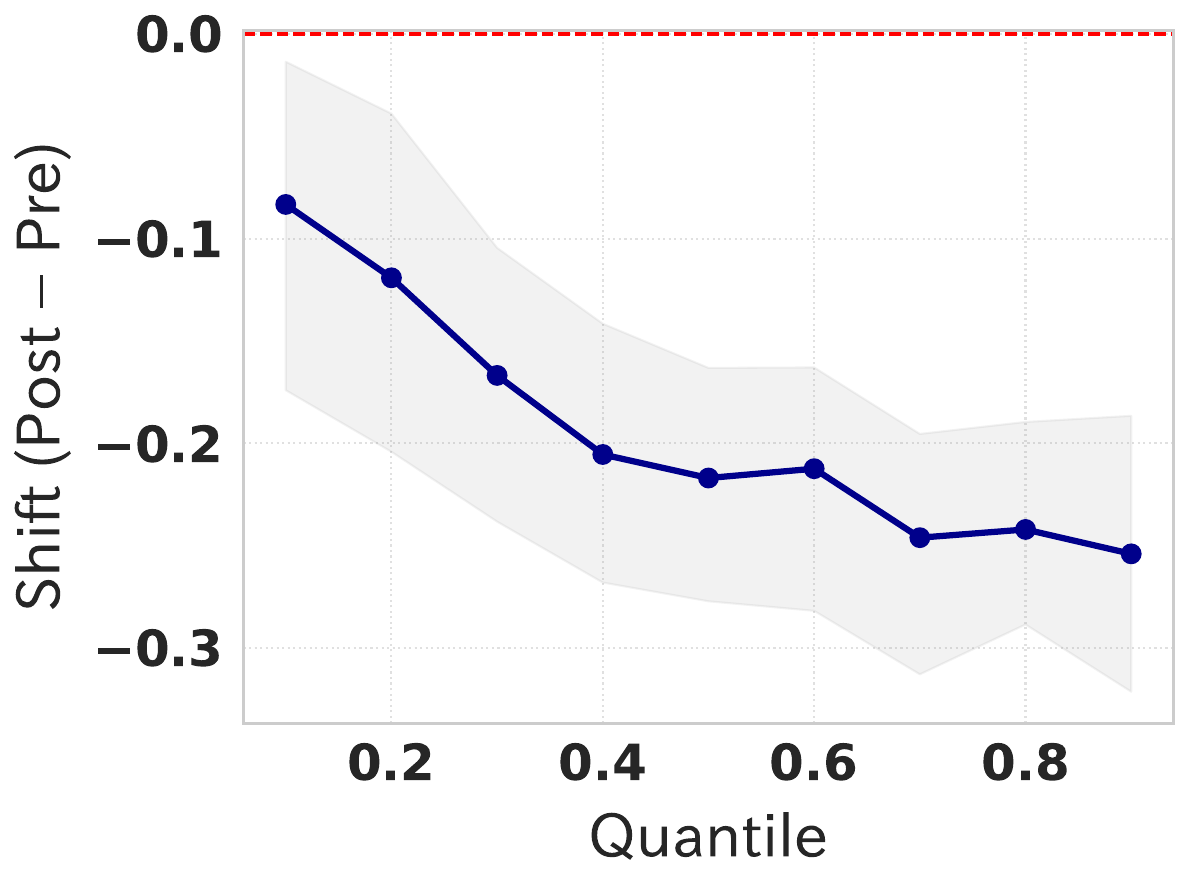}}%
    \hfill
    \subfloat[Boilerplate\label{fig:shift_boilerplate}]{\includegraphics[width=0.48\linewidth]{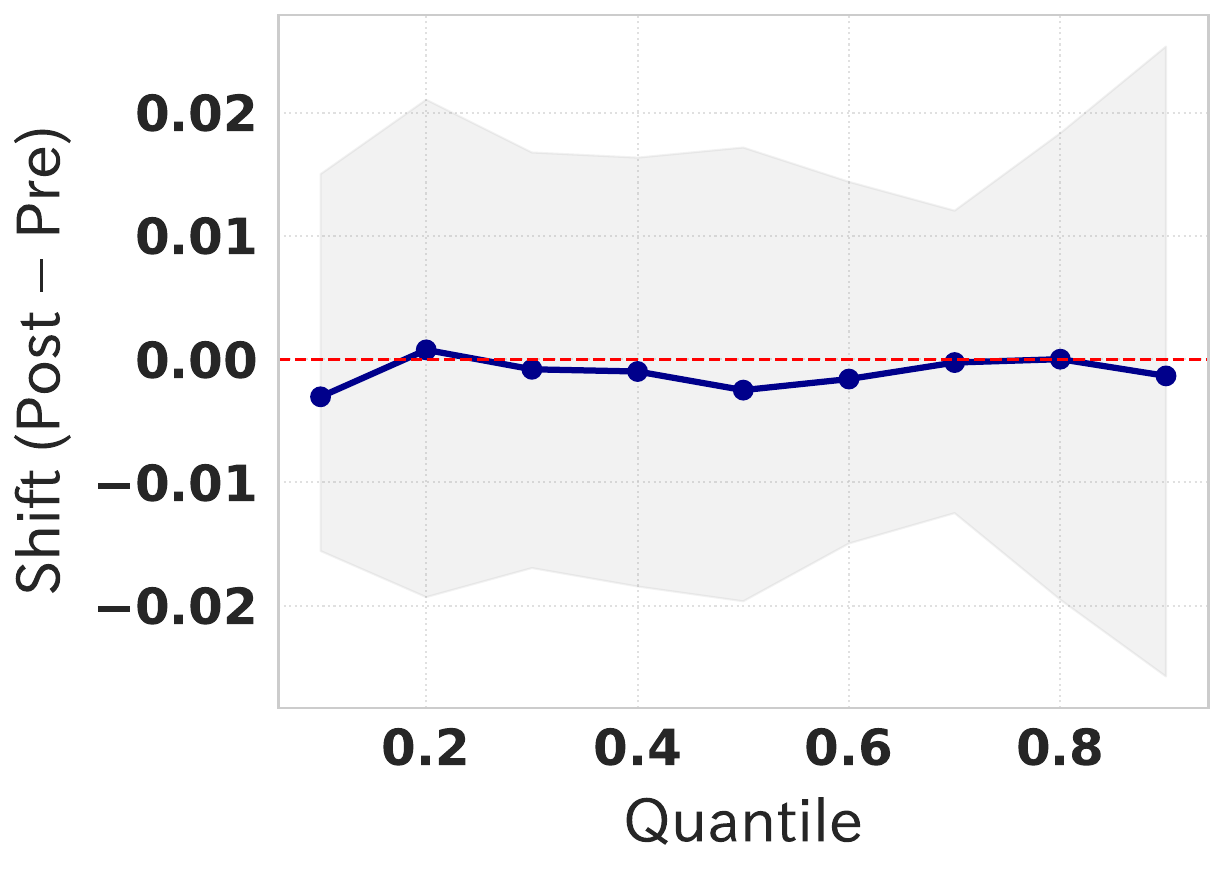}}\\
    \subfloat[Stickiness\label{fig:shift_stickiness}]{\includegraphics[width=0.48\linewidth]{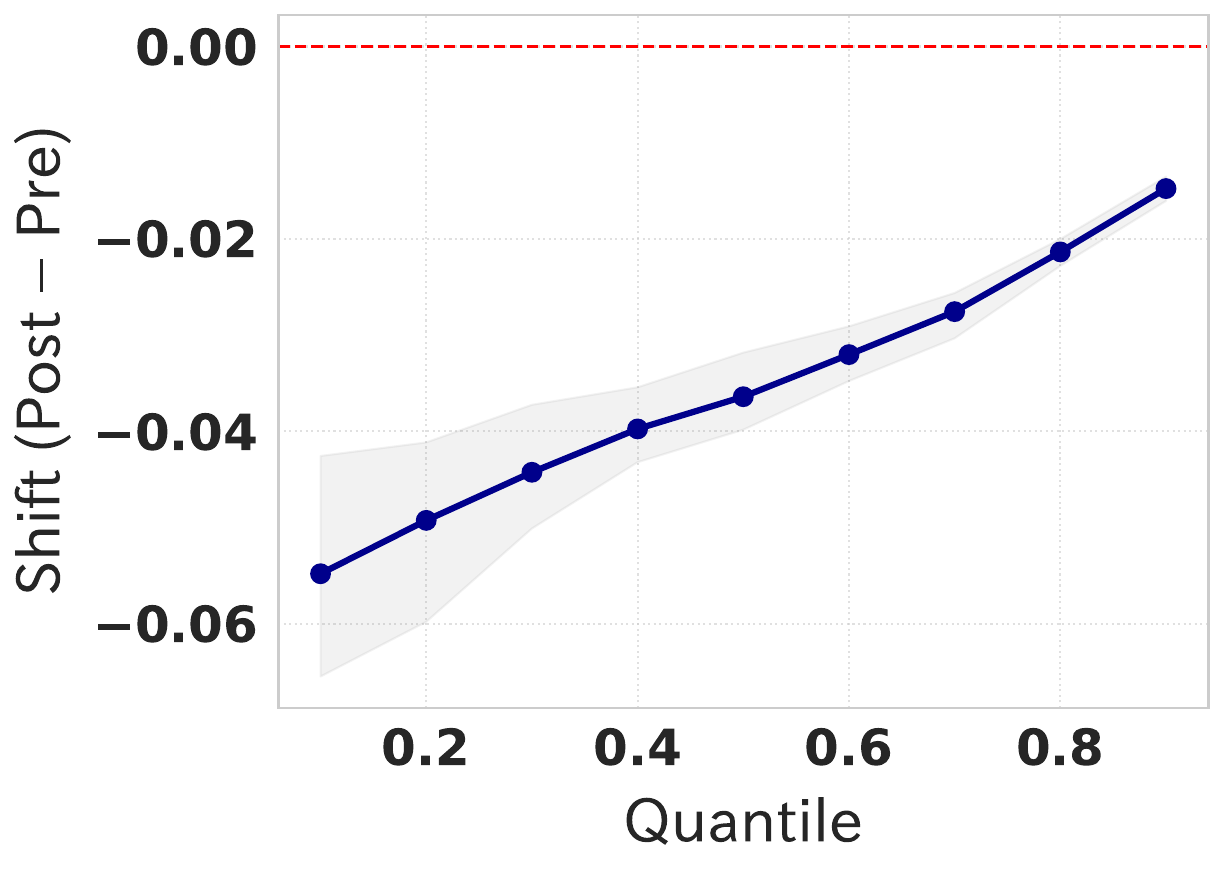}}%
    \hfill
    \subfloat[Relevance\label{fig:shift_relevance}]{\includegraphics[width=0.48\linewidth]{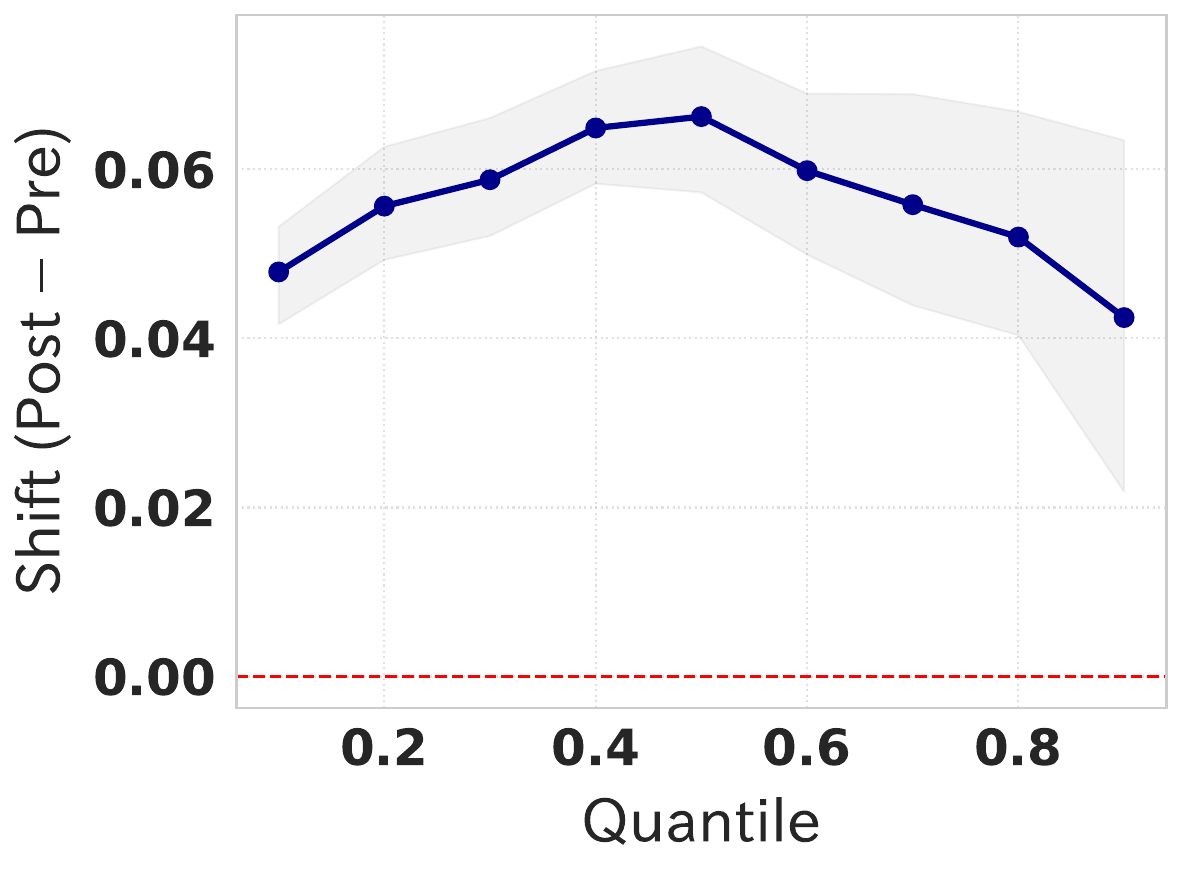}}
    \caption{Shift functions of quantification metrics (Post--Pre).
    Shaded areas show 95\% bootstrap confidence intervals.
    The shift functions show a contrast between heterogeneous and broad-based firm responses: while the volume expansion and specificity decline were largely driven by specific quantiles (lower and upper deciles, respectively), the improvements in structural metrics (stickiness and relevance) and the deterioration in readability occurred systematically across all firm quantiles.}
    \label{fig:shift}
  \end{minipage}
\end{figure*}

\section{Empirical Application}

To demonstrate the utility of our multidimensional framework, we apply it to evaluate Japan's 2019 corporate risk disclosure reforms.

\subsection{Research Questions}

We structure our empirical application around four research questions (RQs), each addressing a distinct facet of the regulatory context:

\textbf{RQ1} (Quantitative Enhancement): Did disclosure volume increase significantly during the post-reform period?

\textbf{RQ2} (Descriptive Quality): Did the disclosures become more specific and readable, or did they remain formulaic?

\textbf{RQ3} (Information Structure): Are firms actively updating their risk information and integrating it more closely with overall management strategy?

\textbf{RQ4} (Market Heterogeneity): Does the magnitude of this disclosure transformation vary systematically across different market segments (e.g., Prime vs.\ Growth)?

\subsection{Data}

Our analysis spans a 10-year period from FY2015 to FY2024 (fiscal years ending in March).
We restrict our sample to companies listed on Japanese stock exchanges that provide continuous narrative disclosures throughout the decade.
Data were retrieved from the Financial Services Agency's EDINET system, yielding a balanced panel of 19{,}770 firm-year observations across 1{,}977 companies.

Risk information and management strategy sections were extracted from EDINET's XBRL-formatted annual securities reports using their standardized element IDs for the "Business Risks" and "Business Policy, Business Environment, and Issues to Address" sections, respectively.

We exclude FY2019 from the pre- and post-reform comparison windows, as it represents the initial year of reform implementation during which one-time compliance adjustments may conflate reform effects with sustained behavioral changes.
FY2019 is nonetheless retained in the time-series visualization to illustrate the timing of the structural break.

\subsection{Results}

This subsection presents the empirical findings, systematically addressing each of the four RQs to uncover the multidimensional dynamics of the disclosure reform.

\subsubsection{Quantitative Enhancement (RQ1)}

As detailed in Table~\ref{tab:results}, disclosure volume expanded significantly from a pre-reform mean of 7.655 to a post-reform mean of 8.145.
The time-series trend in Fig.~\ref{fig:timeseries}(a) illustrates a sharp, sustained upward shift beginning in FY2019, suggesting that higher baseline volumes persisted throughout the post-reform period.
Furthermore, the shift function in Fig.~\ref{fig:shift}(a) reveals that this expansion was particularly pronounced among lower-decile firms, those that had previously provided minimal risk disclosures.
This finding highlights a key advantage of our approach: beyond confirming a mean-level increase, the shift function isolates exactly where in the firm distribution the increase was most pronounced.

\begin{table*}[tp]
\centering
\caption{Correlation Matrix of Rates of Change ($\Delta$).}
\label{tab:correlation}
\begin{tabular}{lrrrrrr}
\toprule
Variables & \multicolumn{1}{c}{(1)} & \multicolumn{1}{c}{(2)} & \multicolumn{1}{c}{(3)} & \multicolumn{1}{c}{(4)} & \multicolumn{1}{c}{(5)} & \multicolumn{1}{c}{(6)} \\
\midrule
(1) $\Delta$ Volume      & 1.000\phantom{$^{***}$} & & & & & \\
(2) $\Delta$ Specificity & 0.182$^{***}$            & 1.000\phantom{$^{***}$} & & & & \\
(3) $\Delta$ Readability & $-$0.222$^{***}$          & $-$0.060$^{***}$         & 1.000\phantom{$^{***}$} & & & \\
(4) $\Delta$ Boilerplate & $-$0.005\phantom{$^{***}$} & 0.029\phantom{$^{***}$} & $-$0.022\phantom{$^{***}$} & 1.000\phantom{$^{***}$} & & \\
(5) $\Delta$ Stickiness  & $-$0.160$^{***}$           & $-$0.141$^{***}$          & $-$0.118$^{***}$           & 0.030\phantom{$^{***}$}  & 1.000\phantom{$^{***}$} & \\
(6) $\Delta$ Relevance   & 0.370$^{***}$              & 0.082$^{***}$             & $-$0.152$^{***}$           & $-$0.012\phantom{$^{***}$} & $-$0.096$^{***}$ & 1.000\phantom{$^{***}$} \\
\bottomrule
\end{tabular}
\begin{flushleft}
\footnotesize
\textbf{Note:} The matrix identifies the relationship between changes in different metrics, including a negative correlation between changes in volume and readability.
$^{***}p < 0.01$, $^{**}p < 0.05$, $^{*}p < 0.1$.
\end{flushleft}
\end{table*}
\subsubsection{Qualitative Transformation: Descriptive Characteristics (RQ2)}

In stark contrast to the quantitative expansion, descriptive quality metrics exhibited a widespread deterioration.
As Table~\ref{tab:results} shows, specificity declined significantly following the reforms. 
The shift function in Fig.~\ref{fig:shift}(b) further reveals that the steepest drops occurred among upper-quantile firms that historically provided highly detailed disclosures.
Similarly, readability worsened across the board. As Fig.~\ref{fig:shift}(c) illustrates, the shift toward greater linguistic complexity occurred systematically across all deciles.

Crucially, as the inter-metric correlation matrix in Table~\ref{tab:correlation} reveals, there is a negative relationship between the rates of change in volume and readability.
This exposes our first major finding: a \textit{volume--readability trade-off}.
Firms broadly added more text during the post-reform period, but this quantitative expansion was accompanied by lower readability. This dynamic is entirely masked when evaluating metrics in isolation.
Meanwhile, the reliance on boilerplate expressions remained virtually unchanged. 
This suggests that principles-based regulation alone may be insufficient to disrupt entrenched, formulaic drafting practices.

\begin{figure*}[tp]
\centering \includegraphics[width=0.85\linewidth]{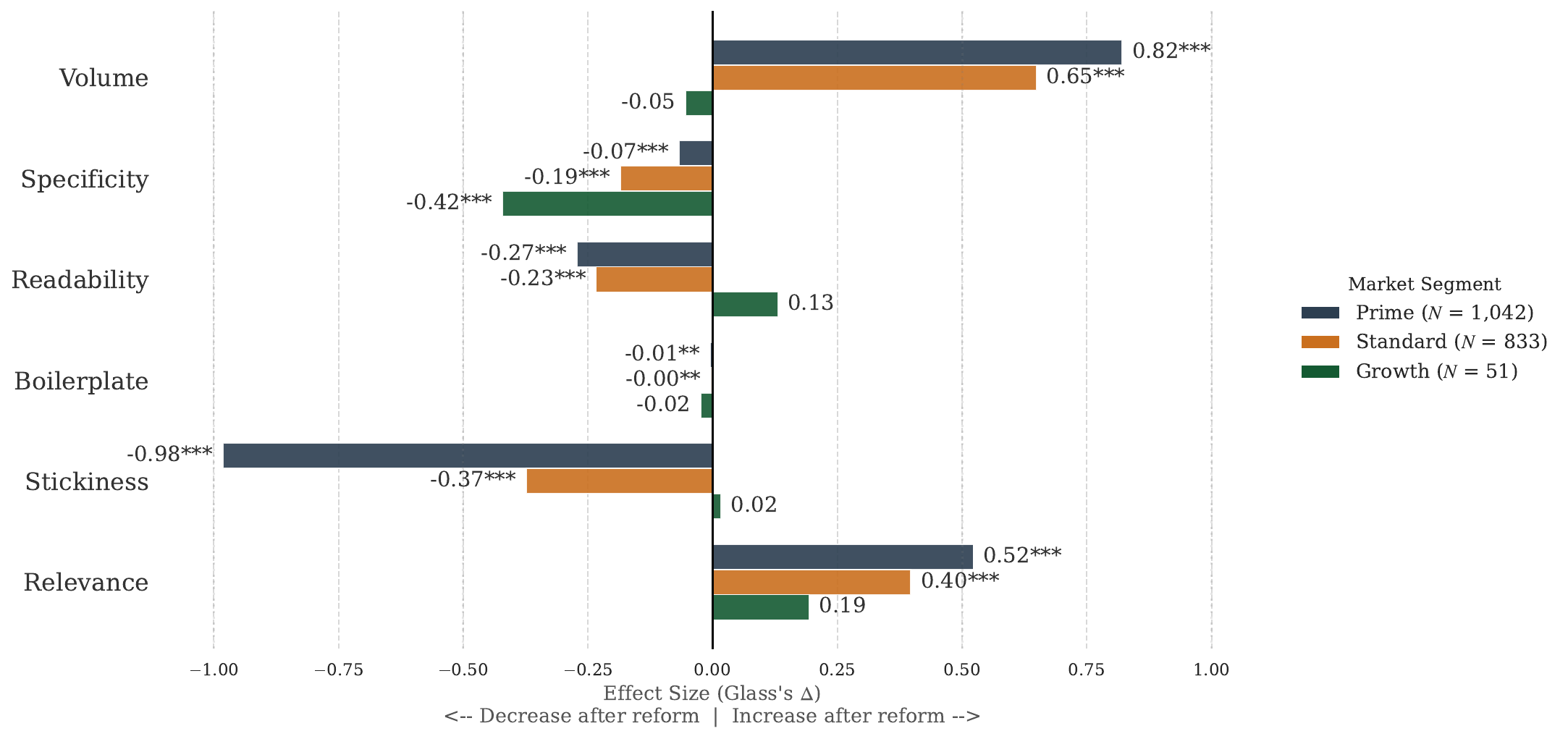} \caption{Changes Following Institutional Reform by Market Segment (Glass's $\Delta$).
Significance levels ($^{***}p < 0.01$, $^{**}p < 0.05$, $^{*}p < 0.1$) are based on paired $t$-tests for mean differences.
The chart shows that the largest effect sizes for volume and relevance are observed in Prime market firms, whereas Growth market firms show limited statistically significant change.}
  \label{fig:effect_size_market}
\end{figure*}

\subsubsection{Qualitative Transformation: Information Structure (RQ3)}

While descriptive quality stagnated, the structural organization of the disclosures underwent profound changes. 
As Table~\ref{tab:results} shows, stickiness declined substantially from 0.957 to 0.922, while relevance increased from 0.213 to 0.268, both representing among the largest effect sizes observed across all six metrics.
Fig.~\ref{fig:timeseries}(e) illustrates a sharp drop in stickiness exactly coinciding with the FY2019 reforms, suggesting that companies undertook substantial, structural rewrites of their risk narratives rather than simply carrying over prior-year text.
Concurrently, as Fig.~\ref{fig:shift}(f) shows, cross-section relevance exhibited a significant upward shift across all firm quantiles, indicating tighter topical alignment with management strategy.
The positive correlation between changes in volume and relevance suggests that part of the added text may have contributed to stronger strategic linkages.

However, juxtaposing these structural improvements with the descriptive deteriorations from RQ2 reveals our second major finding: a \textit{structural--descriptive asymmetry}.
Firms successfully overhauled the macro-organization of their narratives (updating text and linking risks to strategy) but failed to improve the micro-level descriptive quality, leaving the actual risk descriptions abstract, complex, and formulaic.

\subsubsection{Market Segment Heterogeneity (RQ4)}

Finally, we examine whether these disclosure transformations vary systematically across market segments.
To maintain consistency despite the Tokyo Stock Exchange's April 2022 restructuring, we retroactively apply each firm's March 2025 segment classification to the entire study period.
Excluding firms listed solely on regional exchanges yields a subsample of 1{,}042 Prime, 833 Standard, and 51 Growth market companies.
(Note that the smaller Growth market sample limits statistical power).

As illustrated in Fig.~\ref{fig:effect_size_market}, top-tier Prime market companies exhibited the most dramatic effect sizes, largely driving the overall trends of increased volume, decreased stickiness, and heightened strategic relevance.
Conversely, Growth market companies demonstrated virtually no statistically significant changes across any metric except for a decline in specificity.
This disparity constitutes our third major finding: a \textit{market-segment disparity under a uniform regulatory regime}.
The pattern is consistent with greater adjustment difficulties in the Growth segment, suggesting a limitation of applying uniform, principles-based mandates across different market segments.

\section{Discussion}

In this section, we synthesize our empirical findings to discuss the changes in corporate disclosure behavior following the regulatory reforms.
We focus on the inter-metric relationships and heterogeneous responses across market segments, highlighting both the utility and the limitations of our analytical approach.

\subsection{What the Approach Can Reveal}

Our empirical application underscores the fundamental limitation of evaluating narrative disclosures through a single lens.
A conventional analysis focusing solely on volume would erroneously conclude that Japan's 2019 reforms were associated with clear improvements in disclosure quality.
An isolated examination of stickiness or relevance would paint a similarly optimistic picture.
However, by analyzing all six indicators jointly, and by scrutinizing their inter-metric correlations and distributional shifts, our framework exposes underlying qualitative tensions that remain entirely invisible under simpler analytical designs.

\subsection{Key Dynamics Identified by the Approach}

Synthesizing the results across our statistical modules reveals three distinct dynamics in corporate disclosure behavior following the 2019 regulatory reforms.

The \textit{volume–readability trade-off} indicates that quantitative expansion does not necessarily coincide with higher readability.
While disclosure volume rose substantially, readability significantly declined, driven by a negative correlation between their rates of change.
This suggests that, on average, expansions in disclosure were accompanied by declines in linguistic clarity.
Because such inter-metric relationships are invisible to analyses that treat indicators independently, our correlation module was essential for exposing this dynamic.

The \textit{structural--descriptive asymmetry} highlights a critical divergence within the qualitative dimensions themselves.
Information-structure indicators undeniably improved: stickiness decreased as firms actively rewrote text, and relevance to management strategy increased.
Yet, descriptive indicators failed to follow suit, with specificity and readability declining while boilerplate expressions remained entrenched.
The weak correlation between changes in relevance and specificity further suggests that structural reorganizations were not accompanied by more concrete individual descriptions.
Essentially, the post-reform pattern indicates changes in what companies disclose, but less change in \textit{how} they describe it. This finding is consistent with the METI report's assessment~\cite{METI2024} that volume increases have not translated to better communicative effectiveness.

Finally, the \textit{market-segment disparity} exposes the divergence in firm responses across market segments under uniform regulatory mandates.
By stratifying our analysis across market segments, we found that top-tier Prime market companies largely accounted for the overall improvements in information updating and strategic relevance.
By contrast, Growth market firms exhibited virtually no statistically significant changes.
A pooled mean-level comparison would have completely masked this disparity, underscoring the necessity of our segment-level and distributional analyses.

\subsection{Generalizability and Applicability}

A key strength of our analytical framework is its modular design, which facilitates adaptation to other regulatory contexts, disclosure domains, and languages.
Our framework comprises three statistical components: paired testing, shift function analysis, and inter-metric correlation. These are inherently language- and domain-agnostic.
Meanwhile, the six textual indicators can be flexibly swapped or recalibrated depending on the target corpus.

Adapting the framework to non-Japanese disclosures primarily requires substituting the language-dependent modules.
For instance, the Japanese-specific readability formula~\cite{Lee2016} can be replaced with established English counterparts like the Fog Index~\cite{Li2008} or the Flesch--Kincaid metric.
Similarly, the specificity metric relies on named entity recognition, whose coverage must be tuned to the specific NLP toolkit and language used.
Conversely, the stickiness and relevance metrics, based on edit distance and vector similarity, transfer seamlessly, provided robust sentence segmentation is available.

Beyond linguistic adaptation, the framework can be extended to incorporate additional dimensions such as sentiment, forward-looking language, or uncertainty markers without altering the underlying statistical pipeline.
Furthermore, while our current shift function analysis compares marginal distributions, future research could integrate joint-distribution methods or quantile regression to examine how firm-level characteristics modulate disclosure changes.
Applying this approach to other major disclosure regimes, such as the U.S.\ 10-K Risk Factors or EU sustainability reports, represents a natural and promising avenue for future work.

Whether the volume--readability trade-off observed here is universal across disclosure regimes remains an open question.
Dyer et al.~\cite{Dyer2017} document a strikingly similar co-occurrence of length expansion and readability decline in U.S. 10-K filings over 1996--2013.
The fact that a similar pattern emerges across two distinct regulatory regimes suggests that mandatory disclosure requirements may be associated with this trade-off across regulatory contexts.
Future disclosure drafting practices, including the use of generative AI, may further reshape this trade-off.
\subsection{Limitations}

We acknowledge several limitations in both our methodological framework and its empirical application.

Methodologically, our approach is descriptive rather than causal.
While it robustly characterizes multidimensional changes in textual properties, it cannot definitively isolate their underlying causes.
However, our statistical components are compatible with quasi-experimental designs such as difference-in-differences.
Our relevance metric, based on TF-IDF cosine similarity, captures lexical overlap rather than semantic equivalence and may either underestimate alignment when firms paraphrase shared concepts or inflate it when generic boilerplate appears in both sections. 
While the TF-IDF weighting partially mitigates the latter by down-weighting common vocabulary, neural embedding approaches such as Sentence-BERT constitute a natural extension for future work.
Similarly, our Specificity metric depends on NER category choices. GiNZA's `Person' label (0.49\% of total entities) is subsumed under the organizational category, as such mentions predominantly refer to key personnel (e.g., founders, CEOs) constituting organizational risk. Sensitivity analysis across alternative configurations, including Hope et al.'s original four categories and variants excluding business-environment entities, yields firm-level Specificity rankings with Spearman $\rho > 0.99$ for person-inclusion variants, and the post-reform Specificity decline remains significant ($p < 0.001$) across all configurations.

Additionally, our six indicators, while comprehensive, are not exhaustive; other textual dimensions may be highly relevant depending on the research context.
Finally, our shift function analysis compares marginal distributions at matched quantiles and does not fully capture within-firm covariance structures across indicators, an area where joint-distribution analyses could provide deeper insights.

Regarding the empirical application, the post-reform period 
(FY2020--FY2024) overlaps with the macroeconomic disruptions of the 
COVID-19 pandemic. Although the sharp changes in volume and stickiness 
were already evident in FY2019 (Fig.~\ref{fig:timeseries}), which is 
consistent with an early reform-related shift, lingering pandemic 
effects on later years cannot be entirely ruled out.

Furthermore, the limited sample size of the Growth market segment ($N=51$) restricts the statistical power of our inter-segment comparisons.
Lastly, our analysis treats each firm's post-2022 market segment as fixed throughout the 10-year window, utilizing a retroactive assignment rather than explicitly modeling firms' transitions across segments during the Tokyo Stock Exchange restructuring.

\section{Conclusion}

In this study, we developed a longitudinal text analysis approach to capture the multidimensional dynamics of corporate narrative disclosures.
By integrating Japanese-language NLP metric extraction with paired testing, shift function analysis, and inter-metric correlation, our framework mitigates key limitations of single-indicator methodologies.
Crucially, we extended existing indicator sets by introducing a \textit{cross-section relevance} metric to explicitly measure the topical alignment between distinct narrative sections.
Applying this unified approach to a 10-year panel of 19{,}770 firm-year observations surrounding Japan's 2019 disclosure reforms, we identified three recurring patterns in post-reform disclosure change.
Specifically, our analysis revealed a \textit{volume--readability trade-off}, a \textit{structural--descriptive asymmetry}, and a \textit{market-segment disparity}.
Because each of these dynamics was exposed by a distinct component of our statistical toolkit, our findings support the usefulness of multidimensional, distribution-aware research designs.

Beyond the specific context of Japanese regulatory reform, the modular architecture of our approach offers a robust foundation for future disclosure research.
The language-dependent NLP components can be readily substituted for application to English or other linguistic corpora.
Similarly, the statistical layer can be seamlessly expanded to incorporate causal inference methodologies or joint-distribution models.
Applying this framework to other major regulatory shifts, such as the evolution of U.S.\ 10-K Risk Factors or emerging European sustainability mandates, would be a promising direction.
Such comparative studies will be essential for establishing the broader external validity of the disclosure dynamics identified herein.

\bibliographystyle{IEEEtran}
\bibliography{reference}

@ARTICLE{Lee2016,
  title     = "A Study on Text Difficulty for {Japanese} Language Education",
  author    = "Lee, Jae-Ho",
  journal   = "Waseda Studies in Japanese Language Education",
  number    = 21,
  pages     = "1--16",
  year      = 2016,
  note      = "(in Japanese)"
}

@ARTICLE{Nakano2022,
  title     = "Analysis of Narrative Information in Annual Securities Reports: Trends in Information Update Rates",
  author    = "Nakano, Takayuki and Yuasa, Daichi",
  journal   = "Disclosure \& IR",
  volume    = 20,
  pages     = "1--8",
  year      = 2022,
  note      = "(in Japanese)"
}

@ARTICLE{Matsuda2020,
  title     = "{GiNZA} -- Practical {Japanese} {NLP} Based on {Universal Dependencies}",
  author    = "Matsuda, Hiroshi",
  journal   = "Journal of Natural Language Processing",
  publisher = "Association for Natural Language Processing",
  volume    = 27,
  number    = 3,
  pages     = "695--701",
  year      = 2020
}

@ARTICLE{Kuhn1955,
  title     = "The {Hungarian} Method for the Assignment Problem",
  author    = "Kuhn, Harold W.",
  journal   = "Naval Research Logistics Quarterly",
  publisher = "Wiley",
  volume    = 2,
  number    = "1--2",
  pages     = "83--97",
  year      = 1955
}

@ARTICLE{Campbell2014,
  title     = "The Information Content of Mandatory Risk Factor Disclosures in Corporate Filings",
  author    = "Campbell, John L. and Chen, Hsinchun and Dhaliwal, Dan S. and Lu, Hsin-Min and Steele, Logan B.",
  journal   = "Review of Accounting Studies",
  volume    = 19,
  number    = 1,
  pages     = "396--455",
  year      = 2014
}

@ARTICLE{Hope2016,
  title     = "The Benefits of Specific Risk-Factor Disclosures",
  author    = "Hope, Ole-Kristian and Hu, Danqi and Lu, Hai",
  journal   = "Review of Accounting Studies",
  volume    = 21,
  number    = 4,
  pages     = "1005--1045",
  year      = 2016
}

@ARTICLE{Loughran2011,
  title     = "When Is a Liability Not a Liability? {Textual} Analysis, Dictionaries, and 10-{K}s",
  author    = "Loughran, Tim and McDonald, Bill",
  journal   = "Journal of Finance",
  publisher = "Wiley",
  volume    = 66,
  number    = 1,
  pages     = "35--65",
  year      = 2011
}

@ARTICLE{Levenshtein1966,
  title  = "Binary Codes Capable of Correcting Deletions, Insertions, and Reversals",
  author = "Levenshtein, Vladimir I.",
  journal = "Soviet Physics Doklady",
  volume = 10,
  number = 8,
  pages  = "707--710",
  year   = 1966
}

@MISC{METI2024,
  title  = "Study Group on Corporate Information Disclosure: Issues and Future Directions (Interim Report)",
  author = "{Study Group on Corporate Information Disclosure}",
  howpublished = "Ministry of Economy, Trade and Industry",
  year   = 2024,
  note   = "(in Japanese)"
}

@ARTICLE{Dyer2017,
  title     = "The Evolution of 10-{K} Textual Disclosure: {Evidence} from {Latent Dirichlet Allocation}",
  author    = "Dyer, Travis and Lang, Mark and Stice-Lawrence, Lorien",
  journal   = "Journal of Accounting and Economics",
  publisher = "Elsevier",
  volume    = 64,
  number    = "2--3",
  pages     = "221--245",
  year      = 2017
}

@article{carle2023text,
  author    = {Carl{\'e}, Tobias and Pappert, Nicolas and Quick, Reiner},
  title     = {Text Similarity, Boilerplates and Their Determinants in Key Audit Matters Disclosure},
  journal   = {Corporate Ownership \& Control},
  volume    = {20},
  number    = {2},
  pages     = {49--62},
  year      = {2023},
  doi       = {10.22495/cocv20i2art4},
}

@MISC{FSA2019,
  title     = "Principles Regarding Disclosure of Narrative Information",
  author    = "{Financial Services Agency}",
  year      = 2019,
  note      = "(in Japanese)"
}

@ARTICLE{Li2008,
  title     = "Annual Report Readability, Current Earnings, and Earnings Persistence",
  author    = "Li, Feng",
  journal   = "Journal of Accounting and Economics",
  publisher = "Elsevier",
  volume    = 45,
  number    = "2--3",
  pages     = "221--247",
  year      = 2008
}

@ARTICLE{Ito2021,
  title  = "Analysis of {MD\&A}, Risk, and Governance Information Using Text Mining",
  author = "Ito, Takeaki and Kim, Hyonok and Yazawa, Kenichi",
  journal = "Aoyama Management Review",
  volume = 56,
  number = 1,
  pages  = "59--84",
  year   = 2021,
  note   = "(in Japanese)"
}

@MISC{FSA2018,
  title  = "Report of the {Financial Services Agency Disclosure Working Group} -- Toward Realizing a Virtuous Cycle in Capital Markets",
  author = "{Financial Services Agency Disclosure Working Group}",
  year   = 2018,
  note   = "(in Japanese)"
}

@article{Miihkinen2012,
  author    = {Antti Miihkinen},
  title     = {What drives quality of firm risk disclosure? {The} impact of a national disclosure standard and reporting incentives under {IFRS}},
  journal   = {The International Journal of Accounting},
  volume    = {47},
  number    = {4},
  pages     = {437--468},
  year      = {2012},
}

@misc{FSA2019b,
  author = {{Financial Services Agency}},
  title  = {Collection of Good Disclosure Practices for Narrative Information},
  year   = {2019},
  note   = {(in Japanese)}
}

@article{doksum1974,
  author    = {Doksum, Kjell},
  title     = {Empirical Probability Plots and Statistical Inference for Nonlinear Models in the Two-Sample Case},
  journal   = {The Annals of Statistics},
  volume    = {2},
  number    = {2},
  pages     = {267--277},
  year      = {1974},
}

@article{rousselet2017,
  author    = {Rousselet, Guillaume A. and Pernet, Cyril R. and Wilcox, Rand R.},
  title     = {Beyond Differences in Means: Robust Graphical Methods to Compare Two Groups in Neuroscience},
  journal   = {European Journal of Neuroscience},
  volume    = {46},
  number    = {2},
  pages     = {1738--1748},
  year      = {2017},
}

\end{document}